\documentclass[10pt,twocolumn,letterpaper]{article}

\usepackage{cvpr}              

\definecolor{cvprblue}{rgb}{0.21,0.49,0.74}
\usepackage[pagebackref,breaklinks,colorlinks,allcolors=cvprblue]{hyperref}
\usepackage{amsmath,amsfonts,bm}
\usepackage{xcolor,colortbl}
\usepackage{tikz}
\usepackage{arydshln}
\usepackage[accsupp]{axessibility}
\usepackage{comment}
\def\paperID{3979} 
\def\confName{ICCV}
\def\confYear{2025}

\title{ImHead: A Large-scale Implicit Morphable Model for Localized Head Modeling }
\author{Rolandos Alexandros Potamias, Stathis Galanakis, Jiankang Deng \\ Athanasios Papaioannou, Stefanos Zafeiriou
\\ Imperial College London \\ \url{https://rolpotamias.github.io/imHead/}
}

\begin{document}
\definecolor{tabfirst}{RGB}{182,215,168}
\definecolor{tabsecond}{RGB}{207,234,215}
\definecolor{tabthird}{RGB}{255,242,204}
\definecolor{TableRed}{RGB}{244,204,204}

\twocolumn[{
\renewcommand\twocolumn[1][]{#1}%
\maketitle
\begin{center}
    \centering
    \captionsetup{type=figure}
    \includegraphics[width=1\textwidth]{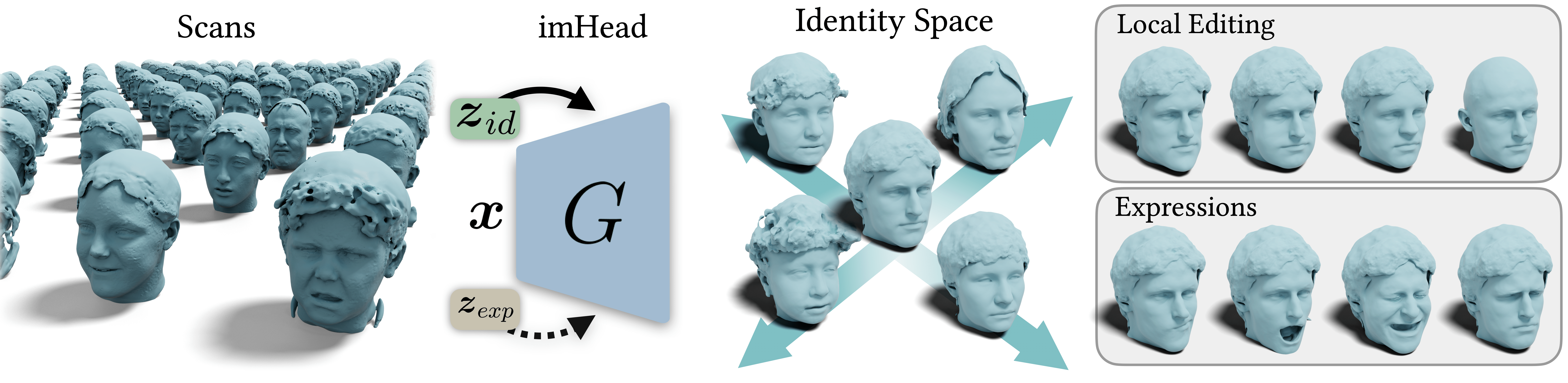}    
    \vspace{-0.4cm}
    \captionof{figure}{We propose \textbf{imHead}, a large scale implicit 3D morphable model composed from 4,000 distinct identities under diverse expressions. imHead enables compact latent representations and localized editing. }
\end{center}}]
\begin{abstract}
Over the last years, 3D morphable models (3DMMs) have emerged as a state-of-the-art methodology for modeling and generating expressive 3D avatars. 
However, given their reliance on a strict topology, along with their linear nature, they struggle to represent complex full-head shapes.  
Following the advent of deep implicit functions, we propose imHead, a novel implicit 3DMM that not only models expressive 3D head avatars but also facilitates localized editing of the facial features. 
Previous methods directly divided the latent space into local components accompanied by an identity encoding to capture the global shape variations, leading to expensive latent sizes. 
In contrast, we retain a single compact identity space and introduce an intermediate region-specific latent representation to enable local edits. 
To train imHead, we curate a large-scale dataset of 4K distinct identities, making a step-towards large scale 3D head modeling. 
Under a series of experiments we demonstrate the expressive power of the proposed model to represent diverse identities and expressions outperforming previous approaches. 
Additionally, the proposed approach provides an interpretable solution for 3D face manipulation, allowing the user to make localized edits. 
Data and models are available on our \href{https://rolpotamias.github.io/imHead/}{project page}.
\end{abstract}    
\section{Introduction}
\label{sec:intro}
In the era of digital avatars and immersive reality, face modeling lies in the core of human modeling, with numerous applications in the context of gaming, graphics, and virtual reality ~\cite{ma2021pixel,baltatzis2024neural,zuo2024signs}. Over the past decades, 3D morphable models (3DMMs) \cite{blanz2023morphable} have revolutionized 3D face modeling. Traditionally, 3D Morphable Models (3DMMs) utilize linear Principal Component Analysis (PCA) to capture the statistical variations of 3D facial geometry in a shared, low-dimensional latent space, enabling efficient data compression and improved generalization capabilities \cite{egger20203d,tarasiou2024locally}.

Despite their wide range of downstreaming applications, from 3D reconstruction \cite{sun2023next3d} to animation \cite{sun2024diffposetalk}, PCA-based models suffer from inherent limitations. 
Firstly, 3DMMs, as linear models, fail to capture complex local variations of the human face, resulting in overly-smooth surfaces that lack high frequency details. 
Although non-linear models have been introduced to enhance the expressivity of 3DMMs \cite{zhou2019dense,Spiral,potamias2020learning,gerogiannis2024animateme,potamias2024shapefusion,gerogiannis2025arc2avatar,ran2024high}, their representations still lack the necessary details required for realistic face modeling. 
Secondly, 3DMMs require consistent topology and precise correspondences across the dataset to effectively capture statistical variations from a shared template.
This can significantly constrain the modeling process, as establishing accurate correspondences between the scans and a unified topology template is a labor-intensive and error-prone task~\cite{egger20203d}, limiting 3DMMs on modeling only the facial regions.

Recent advancements in deep implicit functions have demonstrated great potential in modeling 3D assets. 
These methods employ deep neural networks to estimate the signed distance between any query point $\mathbf{x}$ in 3D space and the surface. 
This continuous representation offers significant advantages compared to voxel grid and mesh representations ~\cite{genova2019learning,deng2021deformed,yenamandra2021i3dmm,zheng2023ilsh}, enabling direct modeling of distributions with minimal alignment requirements ~\cite{park2019deepsdf,palafox2021npms,Zheng_2024_CVPR}. 
Implicit morphable models have been proposed ~\cite{yenamandra2021i3dmm,zheng2022imface,giebenhain2023nphm} to address the geometric constrains of 3DMMs, enabling the modeling of non-rigidly deformable 3D faces.
Implicit representations can facilitate learning of high frequency components, like hair, directly from 3D scans, eliminating the need for dense correspondence and registration steps.
However, current implicit 3DMMs~\cite{zheng2022imface,palafox2021npms,giebenhain2023nphm} model the 3D faces using a global entangled latent space which prohibits localized editing and disentangled manipulations, thus limiting their real-world applications. 
In particular, NPHM~\cite{giebenhain2023nphm}, which is currently the state-of-the-art method for 3D face modeling,  follows a latent space partitioning paradigm to capture more accurately local shape details along with a global identity encoding that captures global shape variations. Nevertheless, this is suboptimal for learning compressed representations as the identity information tends to be captured purely on a single latent vector~\cite{tarasiou2024locally}, limiting the potential editing capabilities of the model. 
Instead, we propose the use of a single compact latent space to effectively capture identity variations and transfer the localized components to an intermediate representation. Such formulation can facilitate seamless shape editing and manipulation, while retaining a compact latent space.

Additionally, current implicit 3DMMs rely on datasets with small identity variations, from limited age and ethnicity groups, which does not adequately capture real-world distribution. 
This highly constrains implicit models from becoming a direct replacement of large-scale 3DMMs that are able to capture large shape variations \cite{booth20163d,ploumpis2019combining}. 
Given the limited identity diversity in publicly available full-head datasets, we propose a head completion strategy to curate an extensive full-head dataset comprising of 4,000 subjects, which presents a 10$\times$ increase compared to prior implicit head models.
By scaling the data used, imHead model makes a step towards modeling the real-world distribution. 

In this paper, we introduce imHead, a deep implicit network 3DMM for face and head modeling. In particular: 
\begin{itemize}
    \item We propose imHead, a large-scale implicit model, that generates realistic 3D heads and expressions, with significant more details compared to 3DMMs. 
    \item We illustrate that imHead, despite being trained only for shape modeling, can naturally achieve localized editing without having to enforce any additional constrains. 
    \item We curate a large full head dataset of 50,000 scans from 4,000 identities. The proposed dataset enables imHead to make a step towards generic head modeling, capturing large shape variations. 
\end{itemize}





\section{Related Work}
\noindent\textbf{3D Morphable Models} are parametric models that enable the generation of new 3D faces by modifying their compact latent representations. 
Blanz and Vetter ~\cite{blanz2023morphable} introduced the first parametric model utilizing principal component analysis (PCA) to learn the statistical shape variations of 3D facial scans. 
Follow-up works, have extended 3DMMs to larger datasets that can effectively capture more diverse shapes ~\cite{booth20163d,li2017learning,papaioannou2022mimicme} and full head models to enforce realistic generations  ~\cite{hu2017avatar,dai20173d,ploumpis2019combining}. 
Building on the success of PCA in accurately capturing data distribution, numerous studies have extended 3DMM techniques to model other parts of the human body, including the full body ~\cite{anguelov2005scape,SMPL,STAR} and hands ~\cite{MANO,Handy,potamias2025wilor,zhang2025hawor}.
Yet, a major challenge with linear 3DMMs is their limited ability to capture high-frequency details, combined with their greedy parameter nature. 
To overcome such limitations, several methods~\cite{ranjan2018generating,Spiral} 
 have proposed to represent 3D meshes as graphs and employ non-linear graph neural network to model 3D human face variation, improving both the efficiency and the details of the modeling. 
Nevertheless, both linear and non-linear 3DMM methods fail to adequately represent finer details and rely on overly smooth cranial meshes.  

\begin{figure*}[!ht]
    \centering
    \includegraphics[width=\textwidth]{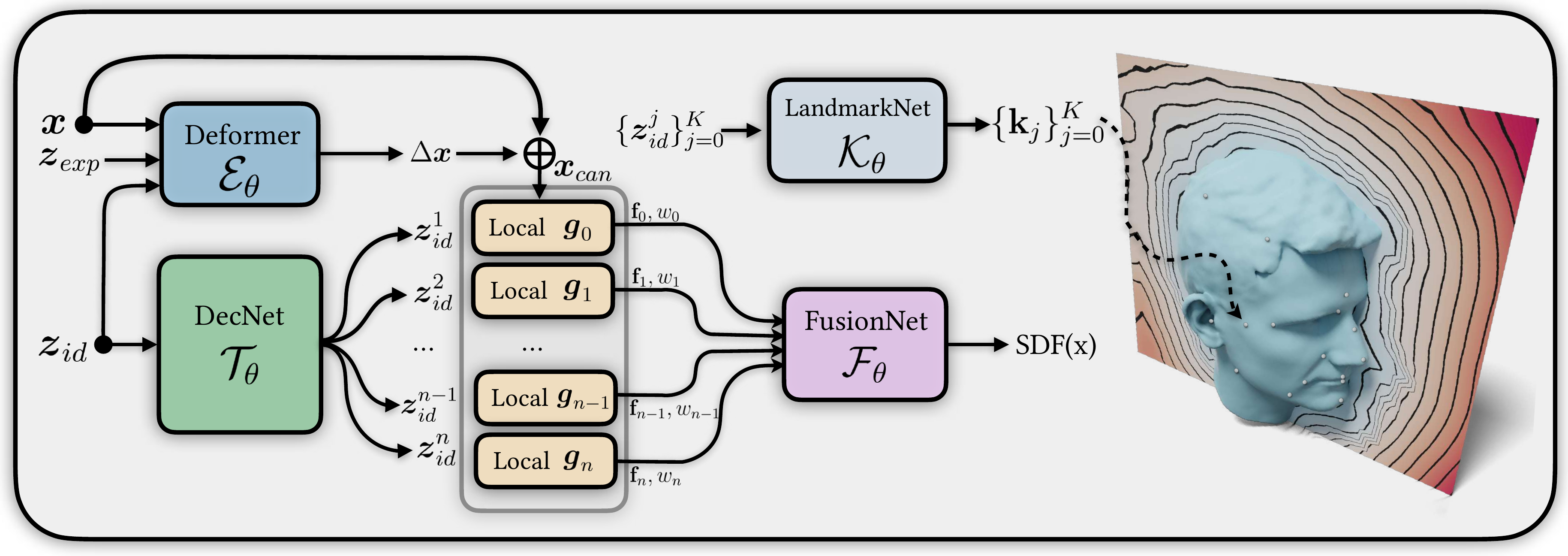}
    \vspace{-0.8cm}
    \captionof{figure}{
    \textbf{Overview of the proposed imHead architecture}: Given a point in the observation space $\bm{x}$ and an expression code $\bm{z}_{exp}$ the \textcolor{CadetBlue}{\textit{Expression Deformer}} network $\mathcal{E}_{\theta}$ predicts a displacement field $\Delta\bm{x}$ to warp the observations to the canonical space $\bm{x}_{can}$. To enable localized editing, \textcolor{ForestGreen}{\textit{DecNet} $\mathcal{T}_{\theta}$} decomposes the global identity latent $\bm{z}_{id}$ into local embeddings $\{\bm{z}^{j}_{id}\}_{j=0}^K$ that correspond to distinct head regions. The local embeddings are used to condition a set of \textcolor{Dandelion}{\textit{Local-Part} $\mathcal{G}_\theta$}  networks that predict localized features $\mathbf{f}_j$ for each point in the canonical space. To facilitate modeling, a landmark regressor \textcolor{gray}{\textit{LandmarkNet} $\mathcal{K}_\theta$} predicts a set of head keypoints,  providing a canonical frame of each local-part network. Finally, the local features are agrregated and fused by \textcolor{violet}{\textit{FusionNet} $\mathcal{F}_\theta$} which regresses the signed distance field of point $\bm{x}$. 
    } 
    \label{fig:method}
\end{figure*}
\noindent\textbf{Deep Implicit Functions (DIFs)} have been well established in the last years given their ability to effectively represent 3D objects of arbitrary topologies. 
In particular, in contrast to explicit methods, such as meshes, implicit functions represent 3D objects and scenes as a continuous function. 
In a pioneering work, Park \etal introduced DeepSDF  ~\cite{park2019deepsdf}, an auto-decoder that models signed distance functions (SDFs) for 3D objects with diverse geometries, demonstrating exceptional performance. 
Genova \etal~\cite{genova2019learning,genova2020local} firstly introduced the notion of localized SDFs and proposed to decompose the global implicit field into local ones, parametrized by 3D Gaussians. 
Deng \etal~\cite{deng2020cvxnet} proposed to model 3D shapes using a collection of local convexes.  
Closer to our work, SPAGHETTI \cite{hertz2022spaghetti} attempted to learn a disentangled representation of 3D objects by introducing an intermediate part-level representation, where 3D Gaussians associated with each part determine the influence and extent of each component.

i3DMM ~\cite{yenamandra2021i3dmm} was the first work that exploited DeepSDF networks to model 3D faces and expressions. 
Given the low resolution of the full-head scans, ImFace ~\cite{zheng2022imface} attempted to learn an implicit function of the frontal face part by introducing a set of local SDF networks that decompose global surface into local geometries. 
To enable training for open surfaces, the authors introduced a pseudo-watertight relaxation. 
Following a similar localized approach, NPHM ~\cite{giebenhain2023nphm} achieved higher generation quality while extending the implicit network to capture the full 3D head.  
However, all of the aforementioned implicit 3DMMs suffer from two key limitations. 
First, they were trained on a limited dataset of fewer than  300 subjects, resulting in small identity variation which limits their potential real-world applications.
In contrast, we scale the size of the training data by 10$\times$ and enable the model to capture a wider range of identity variations.  
Secondly, and more importantly, although these models decompose the 3D head into localized fields, they learn an entangled latent space that prohibits localized face editing and manipulation, a key-feature for real-world applications. 
We propose a simple, yet effective, architecture that provides natural region disentanglement and facilitates smooth manipulation of individual face parts. 

\section{Method}
\subsection{Dataset Curation}
A key factor behind the success of state-of-the-art 3DMMs \cite{booth20163d}, lies in the scale of the training data which adequately captures large variations of the real-world distribution. 
Current implicit methods for 3D head modeling, rely on small datasets with a few hundred of unique identities, limiting their generalization performance to out-of-distribution data. 
This is primarily caused from the scarcity of available large scale 3D head datasets.
To make a step towards large-scale modeling of the human head, we propose an effective pipeline to curate a dataset of over 4,000 distinct identities, that is $10\times$bigger than previous full-head datasets. 
To achieve this, we utilized the raw scans of MimicMe dataset~\cite{papaioannou2022mimicme}, which provides frontal face scans of subjects under 20 different expressions.
We extract 3D landmarks by rendering the scans from multiple viewpoints and applying triangulation to the 2D keypoints detected by an off-the-shelf network \cite{deng2020retinaface}. 
By using iterative closest point (ICP), we rigidly map the facial scans in the FLAME \cite{li2017learning} canonical space and perform a fitting optimization step to estimate some soft correspondences between the scans and the parametric model. 
To handle irregular shapes, we mainly penalize fitting in the face region.
We, then, fit the NPHM model \cite{giebenhain2023nphm} to each scan, minimizing the SDF structural loss \cite{igr} and a 3D landmark loss for five key facial landmarks. In this way, we fill the face scan and acquire the full 3D head model.
Finally, since many of the identity details might have diminished through the fitting process, we perform non-rigid iterative closest point (NICP) registration \cite{amberg2007optimal} between the fitting and the raw scan. 
For additional details about the curated dataset we refer the reader to the supplementary material. 

\subsection{imHead Overview}
We propose an implicit head model $\mathcal{M}$ that given a set of identity $\bm{z}_{id}$ and expression $\bm{z}_{exp}$ latent codes can generate the signed distance field $\mathcal{M}:(\bm{x}, \bm{z}_{id}, \bm{z}_{exp}) \mapsto y \in \mathbb{R}$ of full head 3D avatars in an auto-decoder fashion \cite{bojanowski2017optimizing}. 
The proposed model is founded on three main modules: i) the \textit{identity decomposition network} $\mathcal{T}_{\theta}$ that partitions the global identity encoding $\bm{z}_{id}$ into local shape parts $\bm{z}^{j}_{id}$ ii) the \textit{structure blending network} $\mathcal{F}_{\theta}$ that combines the localized part features and predicts the global implicit field and iii) the backward \textit{expression warping module} $\mathcal{E}_{\theta}$ that learns an observation-to-canonical space mapping to model facial expression deformations. 
Using this formulation, imHead offers two-levels of disentanglement in both expression-identity space and in local-shape canonical space. 
An overview of the proposed model is illustrated in \cref{fig:method}. 

\subsection{Identity-Space Implicit Function}
Our goal is to learn a neural representation of 3D head shapes that enables both global and local shape modeling. 
Previous methods \cite{zheng2022imface,genova2020local} attempt to decompose the 3D canonical space into local parts  
conditioned on a compact global latent code. 
However, despite the achieved latent compression, the expressiveness of such representation not only remains limited but also prohibits localized editing.
Although an obvious solution would be to directly partition the latent space into part-specific latent vectors, this not only diminishes the compactness of the networks, as the latent representation of the shape increases significantly, but also has shown to affect the smoothness of the shape \cite{tarasiou2024locally}.
To avoid such phenomena, NPHM~\cite{giebenhain2023nphm} introduces an additional global identity latent vector which, however, prohibits localized manipulations since the global information is baked in the local networks.   
In contrast, we aim to bridge both worlds and propose an implicit network that utilizes a global latent space $\bm{z}_{id}$ to guide local networks $\mathcal{G}_{\theta} = \{\bm{g}_j\}^K_{j=0}$. 
By using this formulation, we can leverage both compact latent representations and local disentanglement between the different shape parts. 

\noindent\textbf{Decomposition Network (DecNet).} Employing a single global latent code $\bm{z}_{id} \in \mathbb{R}^{d_{g}}$ can enhance both the compactness and the reconstruction performance of the network, since entangled spaces are able to better capture patterns within the data distribution \cite{tarasiou2024locally}.
Aiming to enable localized editing, we utilize a decomposition network $\mathcal{T}_{\theta}$ that maps the entangled global shape representation into $K$ localized part-specific embeddings: 
\begin{equation}
    \{\bm{z}^{j}_{id}\}^K_{j=0} = \mathcal{T}_{\theta}(\bm{z}_{id})
\end{equation}
where $\bm{z}^{j}_{id} \in \mathbb{R}^{d_{l}}$ denotes the $j-$th part-embedding of the $\bm{z}_{id}$ identity. 
Similar to \cite{giebenhain2023nphm}, we partition each face to K=39 local regions defined from a set of corresponding landmark keypoints spanning the head shape. 
We implement $\mathcal{T}_{\theta}$ using a simple linear projection layer. 

\noindent\textbf{Local-Part Networks.} To increase the expressivity of the network and enable localized editing, we divide the modeling workload to $K$ distinct local-part networks $\{\bm{g}_j\}^K_{j=0}$, each of them guided from a corresponding local region embeddings $\bm{z}^{j}_{id}$. 
Each local-part network $\bm{g}_j$ receives a query coordinate $\bm{x} \in \mathbb{R}^3$ along with the part-specific identity embedding and extracts a high dimensional feature $\mathbf{f}^{j}_x$. 
We follow \cite{giebenhain2023nphm} and divide the face into symmetrical and non-symmetrical regions.
This enables us to model the symmetrical regions with a single shared local-part network defined on the left side of the face.
To facilitate the disentangled editing, each local-part network is defined on its own canonical space, centered around its corresponding keypoint $\bm{k}_j$:
\begin{equation}
    \mathbf{f}^{j}_x = \bm{g}_j(\bm{x} - \bm{k}_j, \bm{z}^{j}_{id})
\end{equation}
where $\mathbf{f}^{j}_x$ denotes $j$-th the feature embedding corresponding to point $\bm{x}$ and $\bm{k}_j$ represent the generated landmark keypoint corresponding to region $j$. 
Specifically, to enable end-to-end human head modeling we train a small Landmark-Net MLP $\mathcal{K}$ that regresses the landmark positions $\{ \bm{k}\}_{j=0}^K$ based on the latent encodings $\bm{z}^{j}_{id}$: 
\begin{equation}
     \{ \bm{k}\}_{j=0}^K = \mathcal{K}(\big[||_{j=0}^K \bm{z}^{j}_{id} ])
\end{equation}
where $||$ denotes the concatenation operator. 
We opted to select the latent embeddings $\bm{z}^{j}_{id}$, instead of the global identity latent code $\bm{z}_{id}$, to guide the landmark regression network as it can enable fine-grained and robust keypoints even in manipulated regions. 
Note that query points located in the right facial symmetry regions are first mirrored along the facial symmetry axis to align with the corresponding left region.
To enable modeling of high-frequency surface details we utilize positional encodings \cite{mildenhall2020nerf} to represent region canonical positions $\gamma(\bm{x}-\bm{k}_j)$.

\noindent\textbf{Structure Blending Fusion Network (FusionNet).} In the final step of the proposed identity network, the part-level feature embeddings $\{\mathbf{f}^{j}_x\}^K_{j=0}$ produced from each local-part network are fused to form a global feature embedding that is used to predict the signed distance of query point $\bm{x}$. 
\begin{equation}
    \hat{\mathbf{f}}_x = \sum_{j}^K {w}(\bm{x}, \bm{k}_j) \mathbf{f}^{j}_x
\end{equation}
where ${w}(\bm{x}, \bm{k}_j)$ scales the contribution of each feature embedding based on the position of the point $\bm{x}$: 
\begin{equation}
{{w}}(\bm{x}, \bm{k}_j) = \frac{e^{\frac{-||\bm{x}- \bm{k}_j||_2}{\sigma}}}{\sum_{j}^K e^{\frac{-||\bm{x}- \bm{k}_j||_2}{\sigma}}}
\end{equation}
Finally, the aggregated feature embedding $\hat{\mathbf{f}}_x$ is used to condition the structure network that predicts the signed distance field $y$: 
\begin{equation}
    y = \mathcal{F}_{\theta}(\bm{x}, \hat{\mathbf{f}}_x) \in \mathbb{R}
\end{equation}
Note that, in contrast to previews methods \cite{genova2019learning,zheng2022imface,giebenhain2023nphm}, we do not directly blend the local neural fields as it would result in discontinuities in the global field during the editing process. 
Instead, we rely on a fused feature embedding to guide the global implicit field that facilitates a smooth editing process. 

\subsection{Expression Warping}
To enable animation of the identity space and capture the deformations incurred from facial expressions, we develop a deformation network that aims to learn an observation-to-canonical space mapping. 
In contrast to~\cite{palafox2021npms,giebenhain2023nphm}, that utilize forward deformations, we rely on a backward warping that can facilitate the fitting process and provide a more straight-forward training process. 
In particular, fitting observations using a forward deformation field requires an initial iterative root-finding step to establish soft correspondences between the observation and the canonical space. 
Apart from the additional computation overhead introduced by the root-finding optimization scheme, this approach is highly sensitive to the soft correspondences and even a small error could disrupt the reconstruction process.
Instead, backward warping the observations to the canonical space enables a smooth fitting process similar to traditional 3DMMs. 
In particular, our \textit{Expression Deformer} network $\mathcal{E}$ learns a deformation field to localize the observed posed points $\bm{x}_{obs}$ to the canonical space: 
\begin{equation}
    \Delta \bm{x} = \mathcal{E}\big(\bm{x}_{obs}, \bm{z}_{id},\bm{z}_{exp}\big) \in \mathbb{R}^3
\end{equation}
where $\bm{z}_{id},\bm{z}_{exp}$ denotes the identity and expression latent codes and $\Delta \bm{x}$ the deformation residual. Using this backward warping we can simply derive the point in the canonical space as: 
\begin{equation}
    \bm{x}_{can} = \bm{x}_{obs} + \Delta \bm{x}
\end{equation}
Following ~\cite{giebenhain2024mononphm}, we also predict some additional ambient dimensions $\omega \in \mathbb{R}^2$ \cite{park2021hypernerf} to increase the dynamic capacity of the model. 

\noindent\textbf{Training.}
We train imHead model $\mathcal{M}$ using a combination of loss functions. In particular, we use a set of reconstruction losses, as proposed in ~\cite{igr}: 
\begin{equation}
    \mathcal{L}_{rec} = \sum_{i \in \mathcal{S}}|\mathcal{M}(\bm{x}_i; \bm{z}_{id})|+ ||\nabla_{\bm{x}} \mathcal{M}(\bm{x}_i;\bm{z}_{id}) - \bm{n}_i||  
\end{equation}
that encourage the model $\mathcal{M}$ to vanish on points $\bm{x}_i$ on the ground truth surface $\mathcal{S}$ and their corresponding gradients $\nabla_{\bm{x}} \mathcal{M}$ to match the ground truth surface normals $\bm{n}_i$.
To regularize the gradient values $\nabla_{\bm{x}} \mathcal{M}$ to unit norm space, we also use an \textit{Eikonal term}: 
\begin{equation}
    \mathcal{L}_{eik} =\big(||{\nabla_{\bm{x}} \mathcal{M}(\bm{x};\bm{z}_{id})}|| -1\big)^2 
\end{equation}
Additionally, we supervise the landmark regression network $\mathcal{K}$ using the ground truth landmarks $\hat{\bm{k}}_j$: 
\begin{equation}
    \mathcal{L}_{kpt} = || {\bm{k}}_j - \hat{\bm{k}}_j||_2
\end{equation}
Finally, we regularize the identity and expression latent codes $\bm{z}_{id}$, $\bm{z}_{exp}$ and impose symmetry constraints on the latent embeddings $\bm{z}^j_{id}$ of symmetric regions, similar to~\cite{giebenhain2023nphm} $\mathcal{L}_{sym}$. 
The overall loss function is defined as: 
\begin{equation}
    \label{eq:losses}
    \mathcal{L} = \mathcal{L}_{rec} + \mathcal{L}_{eik} + \lambda_{kpt}\mathcal{L}_{kpt} + \lambda_{sym}\mathcal{L}_{sym} + \lambda_{reg}\mathcal{L}_{reg}
\end{equation}
where $\lambda_{kpt}, \lambda_{sym}, \lambda_{reg}$ are weights to ensure balanced training. 
For additional details regarding the implementation details of our methods, we refer the interested reader to the supplementary material.

\section{Experiments}
In this section we quantitatively and qualitatively evaluate the performance of imHead in reconstruction, generation and editing tasks. 

\noindent\textbf{Baselines.} We compare the proposed method against both implicit and explicit 3D morphable models. Specifically, we evaluate the reconstruction performance of the proposed model against the recent implicit head models monoNPHM~\cite{giebenhain2024mononphm}, NPHM~\cite{giebenhain2023nphm}, NPM~\cite{giebenhain2023nphm} and imFace~\cite{zheng2022imface} along with state-of-the-art explicit BFM~\cite{paysan20093d}, FLAME~\cite{li2017learning}, LSFM~\cite{booth20163d} and a simple PCA model trained on the FLAME fittings. 
Additionally, we compare the editing properties of imHead against NPHM~\cite{giebenhain2023nphm} that shares a local latent space.

\noindent\textbf{Datasets.} To evaluate the proposed and the baseline models we define a test set from MimicMe~\cite{papaioannou2022mimicme} dataset that contains 250 distinct identities which can effectively capture the generalization of each method. We additionally report the reconstruction performance of the competing methods on the test set of  NPHM~\cite{giebenhain2023nphm} dataset which contains 23 identities. 

\subsection{Identity Reconstruction}
We evaluate the identity reconstruction performance of the proposed and the baseline method on MimicMe and NPHM datasets. 
Given that NPHM~\cite{giebenhain2023nphm} and NPM~\cite{palafox2021npms} use an open mouth canonical space, we retrain NPM and NPHM models using a neutral expression canonical space to facilitate a fair evaluation in both MimicMe and NPHM datasets.
In \cref{tab:quantitative_identity}, we report the Chamfer distance (CD) between the ground truth scans and the fittings in the facial region and F-score at 5mm (F@5mm) along with the normal consistency (NC) of the fittings. 
\begin{table}[!h]
  \resizebox{\columnwidth}{!}{
  \centering
  \small
\begin{tabular}{@{}l|ccc|ccc@{}}
& \multicolumn{3}{c}{\textbf{NPHM}} & \multicolumn{3}{c}{\textbf{MimicMe}}    \\ 
\toprule 
{\footnotesize Method} & {\footnotesize CD $\downarrow$} &{\footnotesize NC $\uparrow$} & {\footnotesize F@5mm $\uparrow$} & {\footnotesize CD $\downarrow$} &{\footnotesize NC $\uparrow$} & {\footnotesize F@5mm $\uparrow$} \\ 
\midrule
BFM~\cite{paysan20093d}      & 2.868 & 0.946 &  0.467 &   2.794 & 0.911 &  0.432 \\
LSFM~\cite{booth20163d}     & 1.352 & 0.960 & 0.502  &   1.231 & 0.958 & 0.564 \\
PCA~\cite{blanz2023morphable}      & 1.445 & 0.958 & 0.521 & 1.621 & 0.922 & 0.497 \\
FLAME~\cite{li2017learning}    & 1.244 & 0.943 & 0.632  & 1.336 & 0.929 & 0.606   \\
imFace~\cite{zheng2022imface}   & 0.945 & 0.977 & 0.734 & 0.946 & 0.959 & 0.728 \\
\hdashline
NPM~\cite{palafox2021npms}    & 0.718 & 0.972 & 0.776 & 0.734 & 0.951 &  0.746\\
NPM$\dagger$~\cite{palafox2021npms}    & 0.647 & 0.976 & 0.792 & 0.672 & 0.957 &  0.771\\
\hdashline
NPHM~\cite{giebenhain2023nphm} & 0.558  & 0.977 & 0.848 &  0.618 & 0.966 &  0.798\\
NPHM$\dagger$~\cite{giebenhain2023nphm} & 0.514  & 0.980 & 0.866 &  0.598 & 0.967 &  0.827\\
\hdashline
monoNPHM~\cite{giebenhain2024mononphm} & 0.558  & 0.977 & 0.848 &  0.614 & 0.964 &  0.801\\
monoNPHM$\dagger$~\cite{giebenhain2024mononphm} & 0.514  & 0.980 & 0.866 &  0.593 & 0.968 &  0.829\\
\hline
\textbf{imHead-NPHM} & \cellcolor{tabthird} 0.496    & \cellcolor{tabthird} 0.983 & \cellcolor{tabthird} 0.878 &\cellcolor{tabthird}  0.571 & \cellcolor{tabthird} 0.971 &  \cellcolor{tabthird} 0.838 \\
\textbf{imHead-MimicMe} & \cellcolor{tabsecond} 0.484 & \cellcolor{tabsecond} 0.975 & \cellcolor{tabsecond} 0.874 & \cellcolor{tabsecond} 0.546 & \cellcolor{tabsecond}  0.981 &  \cellcolor{tabsecond} 
 0.851 \\
\textbf{imHead-Full$\dagger$} &  \cellcolor{tabfirst} \textbf{0.459} & \cellcolor{tabfirst} \textbf{0.988} & \cellcolor{tabfirst} \textbf{0.898} & \cellcolor{tabfirst} \textbf{0.533} & \cellcolor{tabfirst} \textbf{0.986} & \cellcolor{tabfirst}{\textbf{0.873}}\\ 
\bottomrule
\end{tabular}
}
\vspace{-0.5em}
    \caption{\textbf{Identity Reconstruction Evaluation} of the proposed and baseline methods using single-scan observations in neutral expressions, even in highly deformable regions such as the eyes and the mouth. $\dagger$ Denotes model trained on the full curated dataset.} 
    \vspace{-0.5em}
  \label{tab:quantitative_identity}%
\end{table}%
To extensively demonstrate the effect of both the devised methodology as well as the impact of the large-scale dataset, we report the reconstruction performance of imHead under three different training setups: using the NPHM dataset (\textit{imHead-NPHM}), using the curated MimicMe dataset (\textit{imHead-MimicMe}) as well as a full combined version (\textit{imHead-Full}). 

As can be easily seen, the proposed approach can outperform previous state-of-the-art methods on both datasets, even when only trained with the NPHM dataset. 
The importance of the large-scale dataset can be validated from the performance of imHead when the curated large-scale dataset is included in the training (\textit{imHead-MimicMe}, \textit{imHead-Full}), exhibiting great generalization to out-of-distribution samples. 
In contrast, NPM~\cite{palafox2021npms}, NPHM~\cite{giebenhain2023nphm} and monoNPHM~\cite{giebenhain2024mononphm} methods face a performance drop when tested on in-the-wild data. 
Similarly, imFace~\cite{zheng2022imface}, apart from modeling only the frontal face part, divides the face in 5 key regions that limits its expressivity to capture sufficient facial details. 
Notably, beyond its strong generalization performance, the imHead model achieves a highly compact latent representation, reducing latent size by 8.5$\times$ compared to the monoNPHM model (256 vs. 2176 in monoNPHM) and by 5$\times$ compared to NPHM model. 

\begin{figure}[!ht]
    \centering
    \includegraphics[width=\linewidth]{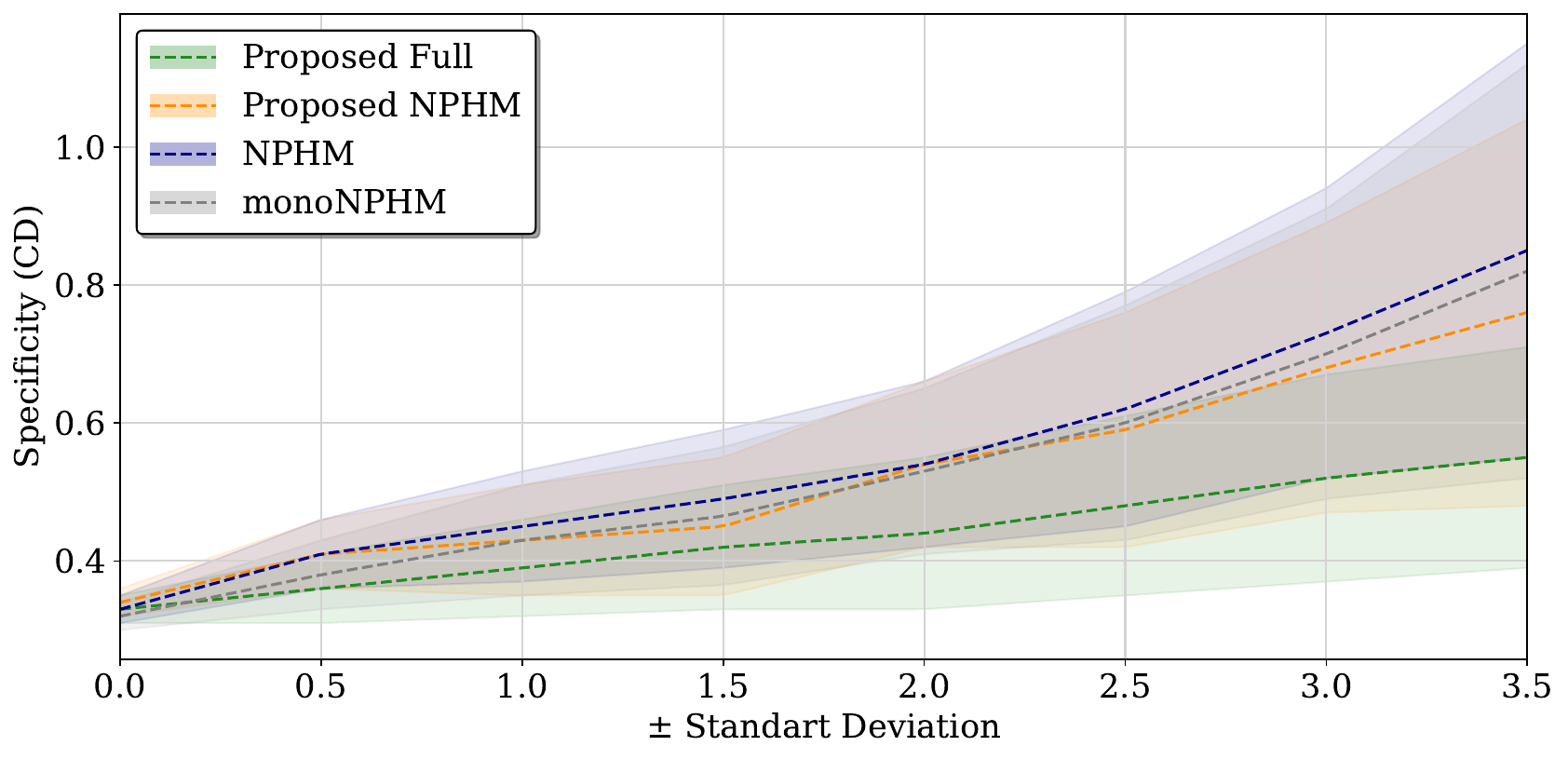}
    \vspace{-0.7cm}
    \captionof{figure}{
    \textbf{Specificity Error} measures the realism of the generated faces under different standard deviation values. 
    } 
    \label{fig:specificity}
\end{figure}

Furthermore, aligned with the evaluation of 3DMMs, we measure the \textit{Specificity} metric which resembles the realism of the face generations that each model produces. 
In particular, we generated 1,000 head meshes from each method and measured their per-vertex distance from the closest real scan sample.
To enable precise sampling from each model, we calculate the statistics of each latent space. 
In \cref{fig:specificity}, we illustrate the specificity error under different standard deviation values. 
The proposed model achieves more stable specificity and scales linearly across the different standard deviation values which indicates that imHead can achieve realistic generation even at extreme latent values. 
\begin{figure}[!h]
    \centering
  \includegraphics[width=\linewidth]{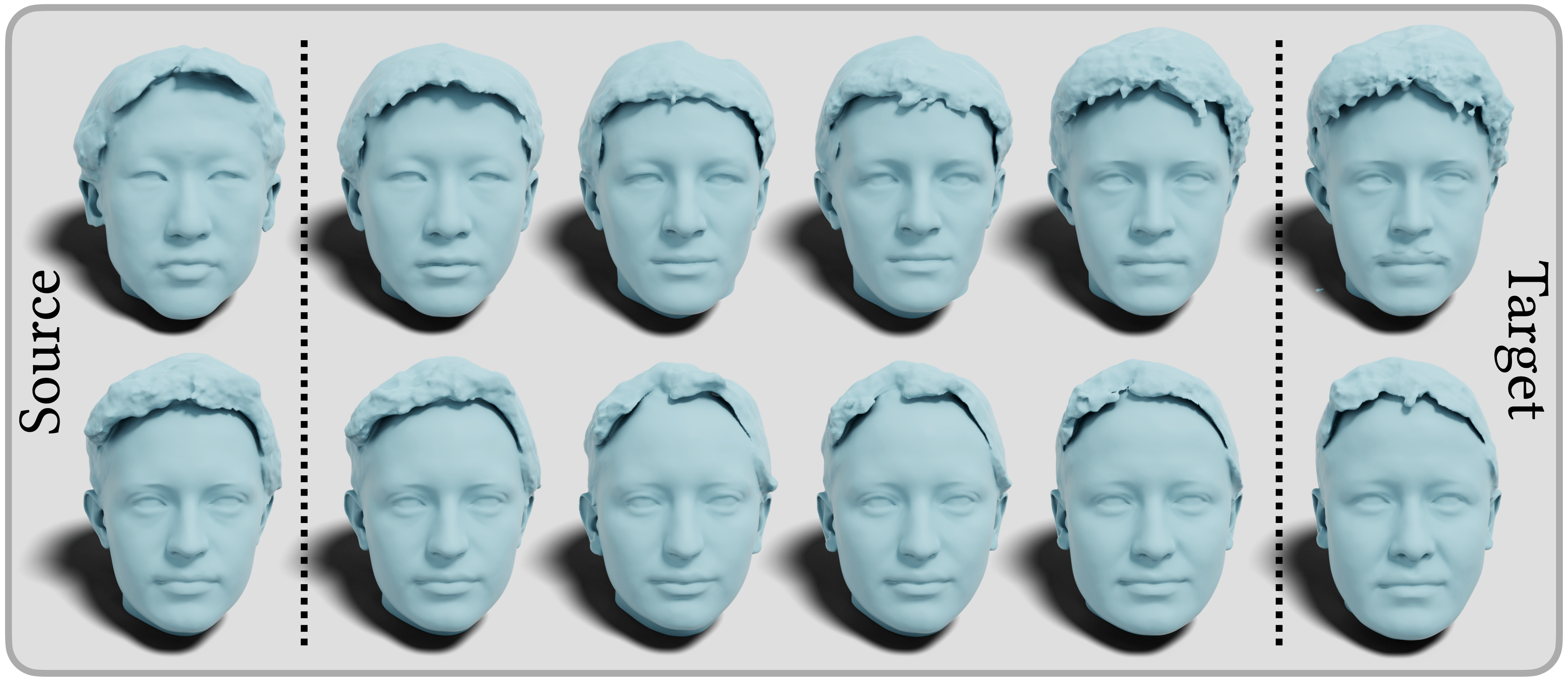}
  \vspace{-0.7cm}
    \captionof{figure}{
    \textbf{Latent Space Interpolation}. The proposed model can achieve smooth changes while interpolating the latent space between source and target identities. } 
    \label{fig:interpolation}
\end{figure}
\subsection{Expression Reconstruction}
Similar to identity reconstruction, to evaluate the expression space of each model we fit a set of test scans with diverse expressions from NPHM and the curated MimicMe datasets. 
For NPM and NPHM models that utilize forward deformations, we use iterative root-finding \cite{chen2021snarf} to fit the expression codes, as suggested in \cite{giebenhain2023nphm}.
We evaluate the reconstructions using the same metrics as in identity reconstruction.
As shown in \cref{tab:quantitative_expression}, imHead can achieve better reconstruction performance compared to previous state-of-the-art methods. 
Similar to the identity space, training the imHead model using only NPHM dataset reduces the model's generalization performance. 
In contrast, when training the model using the curated large-scale dataset, we can achieve more robust reconstructions. 
Please note that using backward deformations, we facilitate the fitting process since our method does not require any iterative root-finding step \cite{chen2021snarf} to map points from the deformed space to the canonical one. 
Instead, our method naturally applies the observation-to-canonical warping through the expression deformer network. 
This results in a huge speed-up in the fitting process as we achieve 3$\times$ faster fitting compared to NPM~\cite{palafox2021npms} and NPHM~\cite{giebenhain2023nphm} models (40sec vs 138sec of NPHM model). 
\begin{table}[!h]
  \resizebox{\columnwidth}{!}{
  \centering
  \small
\begin{tabular}{@{}l|ccc|ccc@{}}
& \multicolumn{3}{c}{\textbf{NPHM}} & \multicolumn{3}{c}{\textbf{MimicMe}}    \\ 
\toprule 
{\footnotesize Method} & {\footnotesize CD $\downarrow$} &{\footnotesize NC $\uparrow$} & {\footnotesize F@5mm $\uparrow$} & {\footnotesize CD $\downarrow$} &{\footnotesize NC $\uparrow$} & {\footnotesize F@5mm $\uparrow$} \\ 
\midrule
BFM~\cite{paysan20093d}      & 2.924 & 0.931 &  0.449 &   2.879 & 0.904 &  0.421 \\
LSFM~\cite{booth20163d}     & 1.396 & 0.954 & 0.497  &  1.307 & 0.951 & 0.553 \\
PCA~\cite{blanz2023morphable}  & 1.463 & 0.953 & 0.512 & 1.672 & 0.910 & 0.599 \\
FLAME~\cite{li2017learning}    & 1.262 & 0.937 & 0.624  & 1.353 & 0.922 & 0.623   \\
imFace~\cite{zheng2022imface}   & 0.966 & 0.971 & 0.756 & 0.987 & 0.945 & 0.742 \\
\hdashline
NPM~\cite{palafox2021npms}     & 0.657	& 0.973 & 0.840     & 0.793 & 0.944 &  0.756\\
NPM$\dagger$~\cite{palafox2021npms}     & 0.648	& 0.975 & 0.837    & 0.729 & 0.948 &  0.774\\
\hdashline 
NPHM~\cite{giebenhain2023nphm}    & 0.526 & 0.976 & 0.892       & 0.679 & 0.959 &  0.798\\
NPHM$\dagger$~\cite{giebenhain2023nphm}    & 0.524 & 0.978	& 0.894      & 0.656 & 0.961 &  0.811\\
\hdashline
monoNPHM~\cite{giebenhain2024mononphm}   & 0.514 & 0.977 & 0.896    & 0.674 & 0.959 &  0.803\\
monoNPHM$\dagger$~\cite{giebenhain2024mononphm}    & \cellcolor{tabthird}0.511 & 0.979	& 0.897      & 0.645 & 0.961 &  0.816\\
\hline
\textbf{imHead-NPHM} & \cellcolor{tabsecond} 0.508 & \cellcolor{tabsecond} 0.980	& \cellcolor{tabthird} 0.898 & \cellcolor{tabthird} 0.623 & \cellcolor{tabthird} 0.963 &  \cellcolor{tabthird} 0.822\\
\textbf{imHead-MimicMe} &  0.513 & \cellcolor{tabthird} 0.979 & \cellcolor{tabsecond} 0.899  & \cellcolor{tabsecond} 0.592 & \cellcolor{tabsecond} 0.968 &  \cellcolor{tabsecond} 0.851 \\ 
\textbf{imHead-Full$\dagger$} & \cellcolor{tabfirst} \textbf{0.485} & \cellcolor{tabfirst}\textbf{0.983} & \cellcolor{tabfirst} \textbf{0.912} & \cellcolor{tabfirst} \textbf{0.563} & \cellcolor{tabfirst} \textbf{0.978} & \cellcolor{tabfirst}{\textbf{0.878}}\\
\bottomrule
\end{tabular}
}
\vspace{-0.35em}
    \caption{\textbf{Expression Reconstruction Evaluation} of the proposed and baseline methods using single-scan observations with diverse expressions. $\dagger$ Denotes model trained on the full curated dataset.} 
    \vspace{-0.3em}
  \label{tab:quantitative_expression}%
\end{table}%

\begin{figure*}[!ht]
    \centering
    \includegraphics[width=0.95\textwidth]{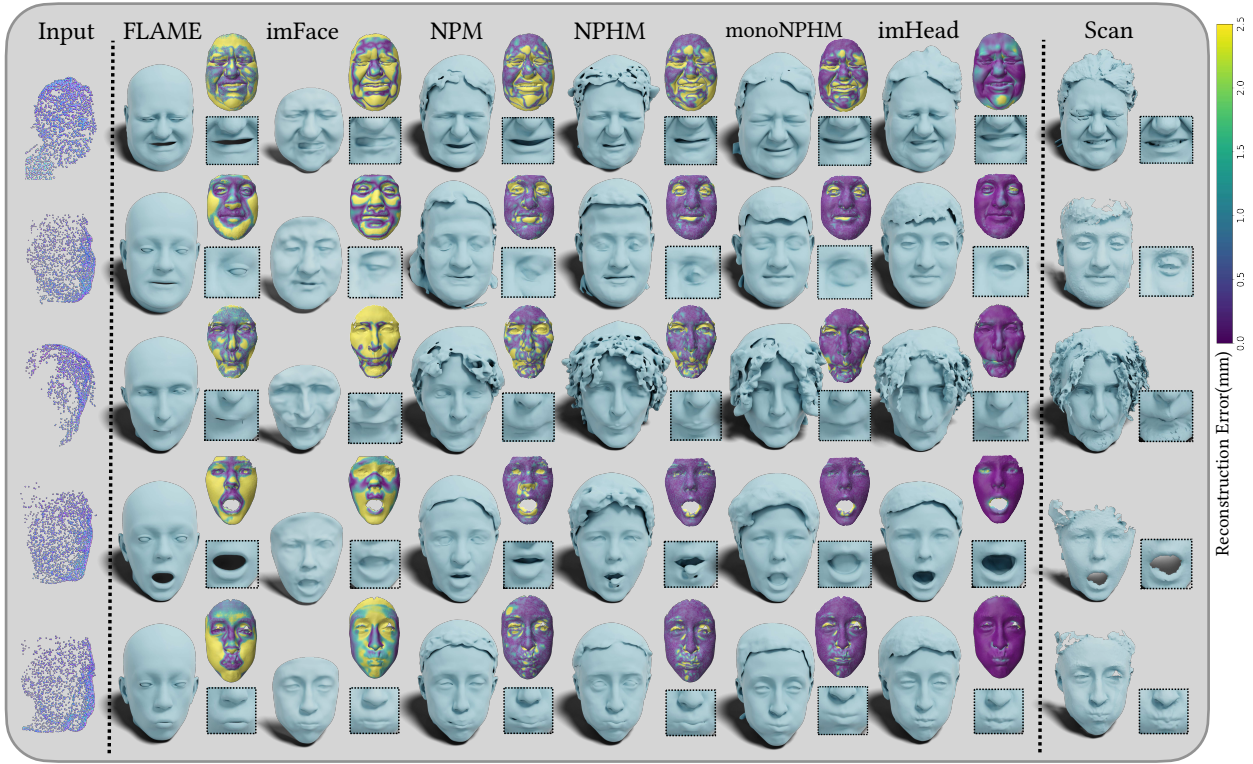}    
    \vspace{-0.35cm}
    \captionof{figure}{
    \textbf{Qualitative Reconstruction Evaluation} of the proposed and the baseline methods under different expressions and identities. Reconstruction for each method is obtained using a fitting optimization from the input partial point clouds. We also report the reconstruction error, in terms of Chamfer distance, color-coded on top of the 3D reconstructions.} 
    \label{fig:method}
\end{figure*}
\subsection{Facial Region Sampling}
\begin{figure*}[!ht]
    \centering
  \includegraphics[width=0.95\linewidth]{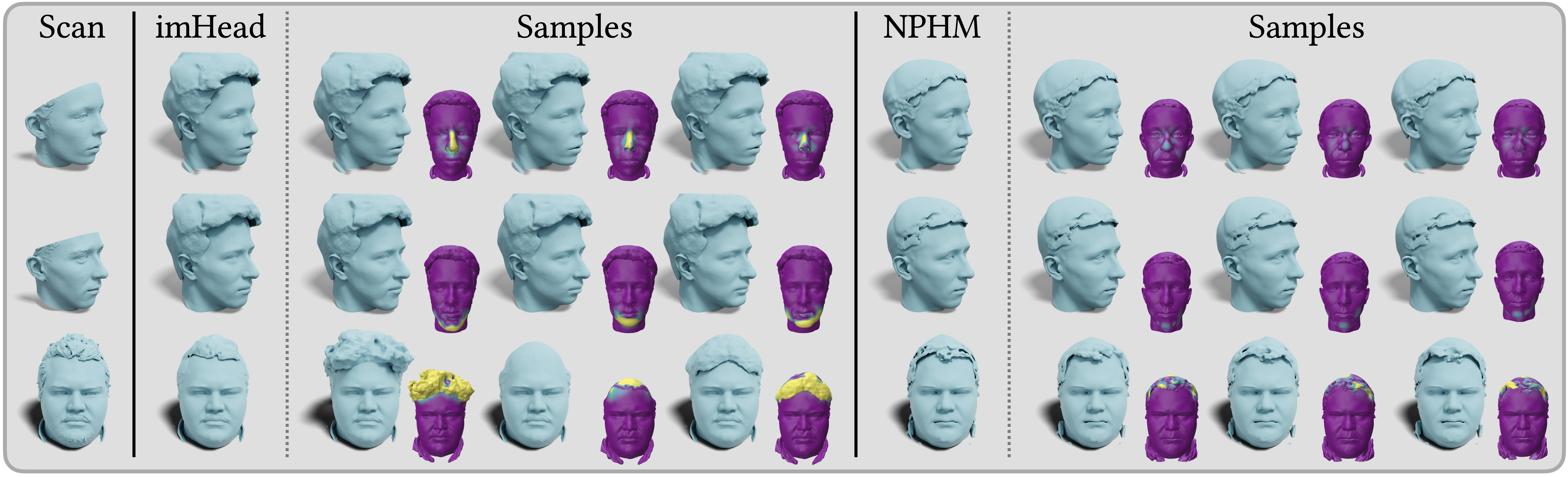}
   \vspace{-0.3cm}
    \captionof{figure}{
    \textbf{Region Sampling.} Given a raw scan (left), we fit imHead and NPHM models to project the scan in the model latent space. We, then, manipulate the region-specific latents by randomly sampling from the latent distribution of each model. We illustrate the displacement changes from the original fitting using color coding. imHead enables more extreme region edits than NPHM, which is limited to region samples within the distribution of the global identity latent space. } 
    \label{fig:region_sampling}
\end{figure*}

A crucial contribution of imHead model is that, by design, it can enable fully localized editing. 
Specifically, the decomposition network (DecNet) allows the proposed model to achieve both global and local identity manipulations. 
To demonstrate the editing capabilities of our model, for a given identity projected in the latent of imHead, we sample a set of local region embeddings $\hat{\bm{z}}_{id}^j$ to substitute the original identity embeddings $\bm{z}_{id}^j$. 
We follow a similar procedure for NPHM~\cite{giebenhain2023nphm} model and modify only the region-specific latent codes. 
In \cref{fig:region_sampling}, we demonstrate the sampled regions for each identity along with a color coded displacement map that quantifies the difference between the original and the modified head. 
It can be easily observed that imHead achieves both realistic and smooth region samples that are fully localized and preserve the rest of the identity unchanged. 
On the contrary, NPHM model is over-constrained from the global identity latent that limits any potential editing capabilities. 

\subsection{Face Part Swapping}
\begin{figure}[!ht]
    \centering
    \includegraphics[width=0.95\linewidth]{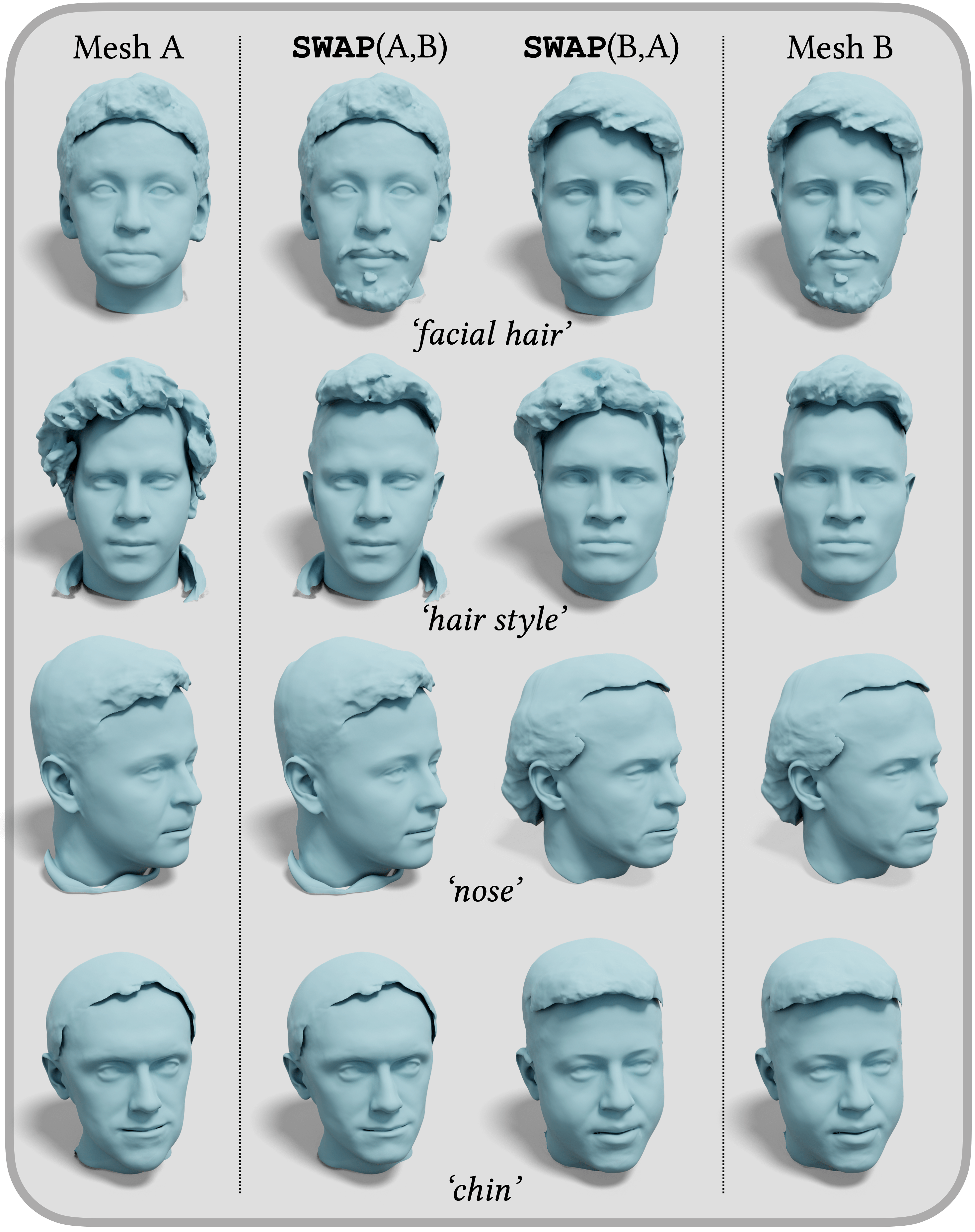}
    \vspace{-0.3cm}
    \captionof{figure}{
    \textbf{Region Swapping.} We visualize the swapping
between the facial regions (from top to bottom: facial hair, hair, nose, and chin) of Mesh A (left) to Mesh B (right)
(SWAP(A,B)) and the opposite (SWAP(B,A)). Note that the changes are fully-localized and do not affect the global identity of each mesh.} 
   \vspace{-0.35cm}
    \label{fig:swaping}
\end{figure}

The localized latent embeddings of the proposed network can facilitate seamless region swapping between different identities. 
In particular, for a given source and target identities, represented in the latent space of imHead, the localized region embeddings enable smooth swapping between source and target features by simply exchanging their local embeddings.
In \cref{fig:swaping} we demonstrate the ability of imHead to swap facial features between source and target identities such as hair, nose, and mouth. 
Note that generated faces preserve the unedited regions and the edits are fully localized without affecting the rest of the identity. 

\subsection{Correspondence Preservation}
\begin{figure}[!ht]
    \centering
    \includegraphics[width=0.95\linewidth]{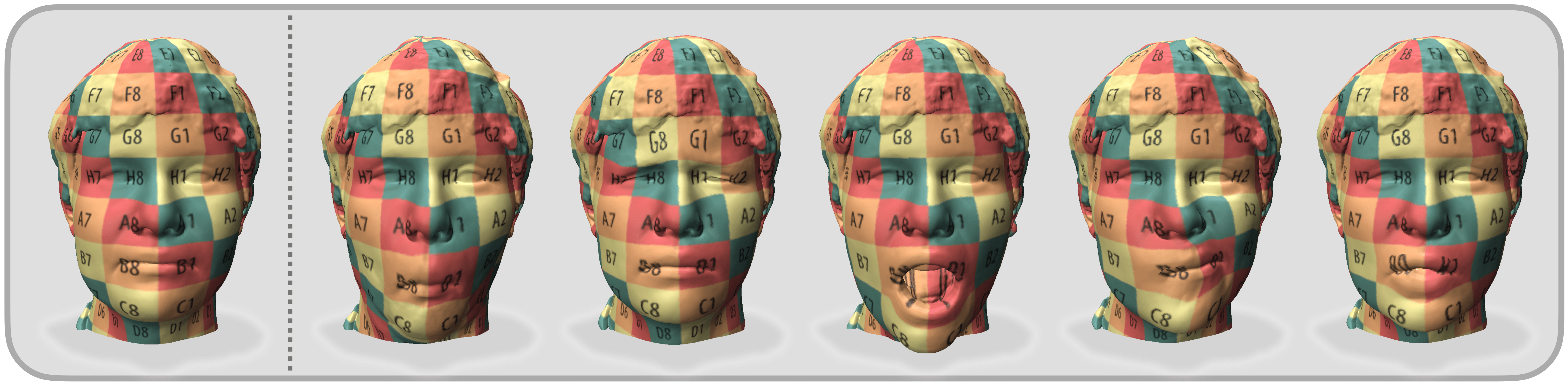}
    \vspace{-0.3cm}
    \captionof{figure}{
    \textbf{Correspondence Preservation} (Left) Source neutral face with UV parametrization applied. (Right) The deformed faces preserve the shape correspondences under different expressions, even in highly deformed regions such as the mouth. } 
    \label{fig:correspondences}
\end{figure}

A key advantage of traditional 3D Morphable Models (3DMM) that rely on a shared template is their ability to maintain dense correspondence across varying expressions.
Preserving the point correspondence is an extremely useful property as it can easily transfer information between different identities and expressions. 
To evaluate the preservation of facial topological semantics, we define a UV map in the canonical space of a mesh by assigning distinct colors to specific vertices and assessing correspondences across different expressions. 
Specifically, we sample a set of diverse expressions and back-project them into the canonical space using the expression deformation network. 
The vertex colors from the original mesh are, then, transferred to the sampled meshes via a nearest-neighbor search (\textit{1-NN}).
Aligned with traditional 3DMMs, imHead implicitly learns a warping that preserves most of the face correspondences in the canonical space.
Specifically, as shown in \cref{fig:correspondences}, imHead UV mapping remains consistent across different expression, even in highly deformed regions such as the mouth. 
\section{Conclusion}
In this work, we introduce imHead, the first large-scale implicit model of the human head, that supports localized face editing, advancing the field of high-fidelity 3D head modeling.
To do so, we curate a large scale dataset that is 10$\times$ bigger than previous full head datasets. 
We highlight the limitations of previous methods in capturing both global and local fields in a stratified manner and propose an effective strategy that enables both compact latent space and localized editing properties.
Under a series of experiments, we demonstrate the superiority of imHead over previous state-of-the-art implicit and explicit 3DMM models, as well as its ability to locally edit 3D heads.

\noindent\textbf{Limitations. }
While imHead tackles several challenges of full-head modeling, it still has certain limitations.
Specifically, imHead shares all the intrinsic limitations of implicit models that struggle to capture high-frequency details, such as hair strands, and suffer from slow inference times compared to explicit 3DMMs.
In addition, although imHead achieves disentangled region modeling, each region depends on multiple nearby anchors to ensure plausible and smooth surfaces, which can slightly affect the desired edits. 
Finally, although a large-scale dataset was curated, it still contains racial biases, including the hair distribution of NPHM dataset, which itself exhibits similar biases.

\noindent\textbf{Acknowledgements.}
S. Zafeiriou and part of the research was funded by the EPSRC Project GNOMON (EP/X011364/1) and Turing AI Fellowship (EP/Z534699/1). R.A. Potamias was supported by EPSRC Project GNOMON (EP/X011364/1).
\maketitle
\setcounter{page}{1}

\section{Ablation Study}
To justify the technical choices we made and evaluate the contribution of each component we perform an ablation study. 

\noindent\textbf{Impact of global latent space.} In particular, we divide the ablation into three major categories to capture all aspects of the proposed model. 
We first evaluate the contribution of the global latent code by modifying imHead latent space to a set of local latents, reported as \textit{w. Local Lat.}.
We follow NPHM and use 32 latent dimensions for each of the K=32 regions resulting in a total 1248 latent space, 4.87$\times$ increase compared to 256 that we use in imHead. 
In addition we report the performance of another variation that extends the local latents to include an additional global identity latent, following the architectural design of NPHM, reported as \textit{w. Local and Global Lat.}. 
The total latent space of this model is 1344 (same as NPHM) which reflects to 5.25$\times$ increase in the latent size. 
Finally, to demonstrate the impact of a single global latent space, we report the results of a model trained with a local latent space where each region receives a local latent of size 8, resulting in a latent space size of 312. 
As can be easily observed in \cref{tab:ablation}, utilizing a split latent space diminishes the reconstruction performance of the network. 
This significantly deteriorates when we use a latent space with the size of 312, where the model struggles to achieve reasonable performance. 
The reason behind this, as suggested in \cite{foti2023led,tarasiou2024locally,potamias2024shapefusion}, is that global patterns of the shape are copied in each local latent which inevitably increase the size of the model. 
To enable a fully local latent space, whilst also achieving sufficient reconstruction performance, it is necessary to increase each latent sufficiently enough to encode both global and local information. 
An intermediate solution is to build a local-global latent space, similar to NPHM model. 
Although this approach achieves similar performance with imHead, it suffers from two main factors: a) a $5\times$ larger latent space which limits the shape compression and b) a highly constrained latent space that prohibits localized face editing as the latent codes are now extended with global information. 
imHead can successfully bridge both worlds by leveraging a compact latent space along with an intermediate localized representation that can facilitated disentangled manipulation. 

\noindent\textbf{Impact of FusionNet.}
To demonstrate the impact of the proposed structural blending network, we train a model that directly regress the local SDF from each local-part network without using an intermediate feature representation as in imHead. 
Despite being slightly lighter model, the performance of the the model drops significantly, as each of the local networks need to directly predict the global SDF. 
It is also important to note that the normal consistency of the reconstructions deteriorates due to non-smooth blending.    
In contrast, when using the proposed FusionNet, the local features are aggregated and the SDF values are regressed using an intermediate feature representation. 
This allows the model to learn more complex representations while achieving smooth reconstructions. 

\noindent\textbf{Impact of Local Canonical Space.}
We additionally report the effect of using a per-region canonical space (\textit{w/o Local Canonical Space}). In particular, each local-part network uses a canonical space that is defined around its corresponding keypoint $\bm{k}_j$ as:
\begin{equation}
    \mathbf{f}^{j}_x = \bm{g}_j(\bm{x} - \bm{k}_j, \bm{z}^{j}_{id})
\end{equation}
where $\mathbf{f}^{j}_x$ denotes the $j$-th feature embedding corresponding to point $\bm{x}$ and $\bm{k}_j$ represent the generated landmark keypoint corresponding to region $j$. 
This canonical space can effectively reduce the workload of each local part network and facilitate the training process. 
As can be seen in \cref{tab:ablation}, apart from the training stability, the canonical space has a positive impact on the reconstruction performance of imHead, as we observe a significant performance improvement when using a canonical space for each local-part network (\textit{imHead-Full}). 
\begin{table}[!h]
  \resizebox{\columnwidth}{!}{
  \centering
  \small
\begin{tabular}{@{}l|ccc|ccc@{}}
& \multicolumn{3}{c}{\textbf{NPHM}} & \multicolumn{3}{c}{\textbf{MimicMe}}    \\ 
\toprule 
{\footnotesize Method} & {\footnotesize CD $\downarrow$} &{\footnotesize NC $\uparrow$} & {\footnotesize F@5mm $\uparrow$} & {\footnotesize CD $\downarrow$} &{\footnotesize NC $\uparrow$} & {\footnotesize F@5mm $\uparrow$} \\ 
\midrule
w. Local Lat. ($d=312$)   & 0.876 & 0.915 & 0.689 & 0.874 & 0.914 &  0.721\\
w. Local Lat. ($d=1248$)  & 0.775 & 0.948 & 0.743 & 0.767 & 0.939 &  0.788\\
w. Local and Global Lat. ($d=1344$) & \cellcolor{tabsecond}0.494 & \cellcolor{tabsecond}0.964 &\cellcolor{tabsecond} 0.841 & \cellcolor{tabsecond}0.569 & \cellcolor{tabsecond}0.958 & \cellcolor{tabsecond}0.857\\
\hline
w/o FusionNet  & \cellcolor{tabthird}0.595 &  \cellcolor{tabthird}0.954 & \cellcolor{tabthird} 0.808 &  \cellcolor{tabthird}0.674 &  \cellcolor{tabthird}0.947 &  \cellcolor{tabthird}0.812\\
w/o Local Canonical Space  & 0.723 & 0.934 & 0.723 & 0.884 & 0.946 &  0.732\\
\hline
\textbf{imHead-Full} &  \cellcolor{tabfirst} \textbf{0.459} & \cellcolor{tabfirst} \textbf{0.988} & \cellcolor{tabfirst} \textbf{0.898} & \cellcolor{tabfirst} \textbf{0.533} & \cellcolor{tabfirst} \textbf{0.986} & \cellcolor{tabfirst}{\textbf{0.873}}\\ 
\bottomrule
\end{tabular}
}
\vspace{-0.5em}
    \caption{\textbf{Ablation Study} of different key components of imHead. } 
    \vspace{-0.5em}
  \label{tab:ablation}%
\end{table}%

\section{Robustness to Noise}
Given that the proposed model was trained on raw scans with a considerable amount of noise, it can achieve robust reconstructions even under noisy point cloud inputs. 
In particular, to evaluate the reconstruction performance of imHead under noise scenarios, we add Gaussian noise of different standard deviations to the input point clouds and measure the performance drop. 
As can be seen in~\cref{fig:noise}, imHead can achieve reasonable reconstruction that retain the identity characteristics even with noise levels that correspond to $1.5$ standard deviations. 
\begin{figure}[!ht]
    \centering
    \includegraphics[width=\linewidth]{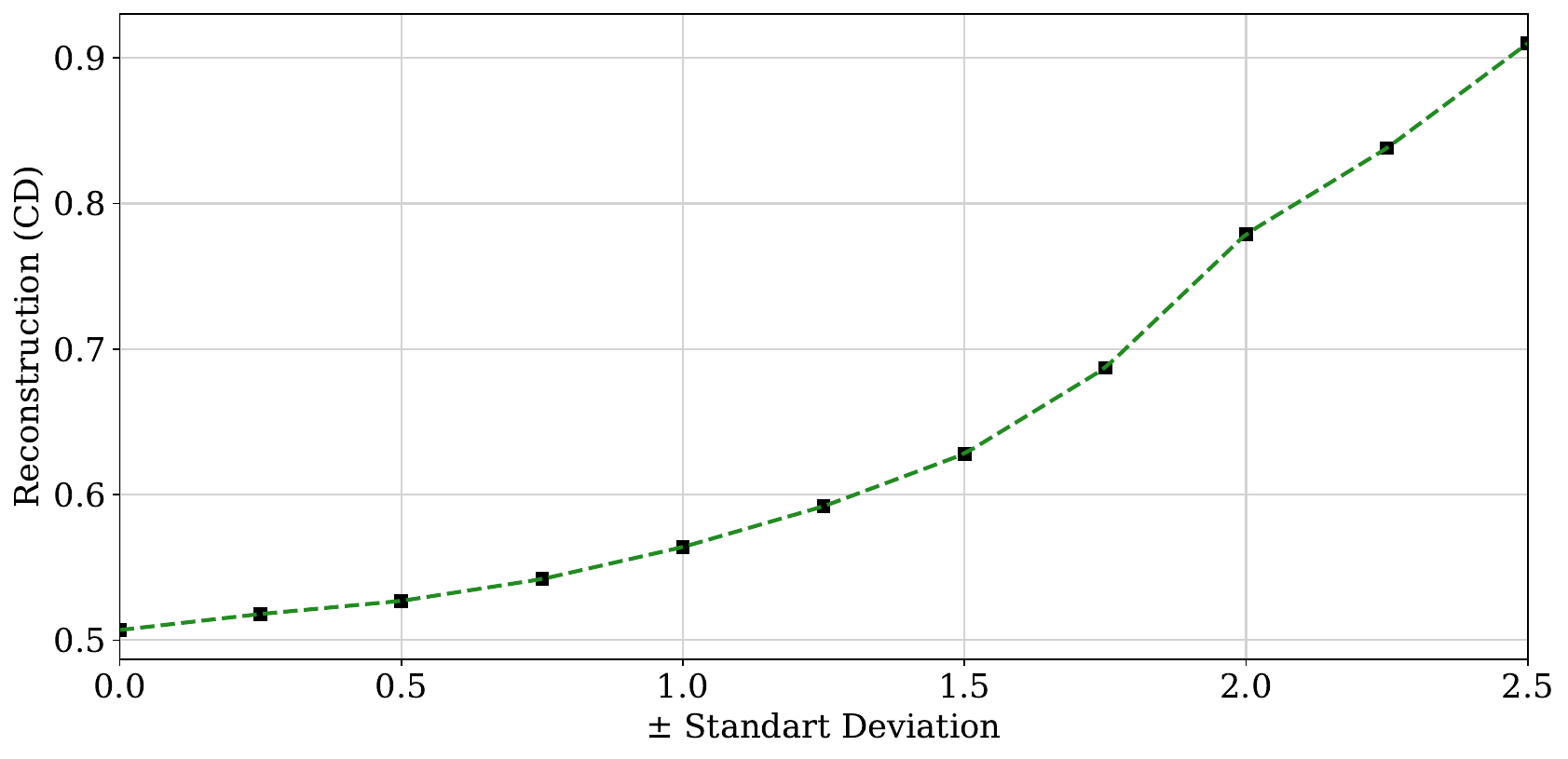}
    \captionof{figure}{
    \textbf{Reconstruction Error under Noisy Inputs}. We measured the reconstruction error under different noise levels of the input point cloud. 
    } 
    \label{fig:noise}
\end{figure}
\begin{figure}[!ht]
    \centering
    \includegraphics[width=\linewidth]{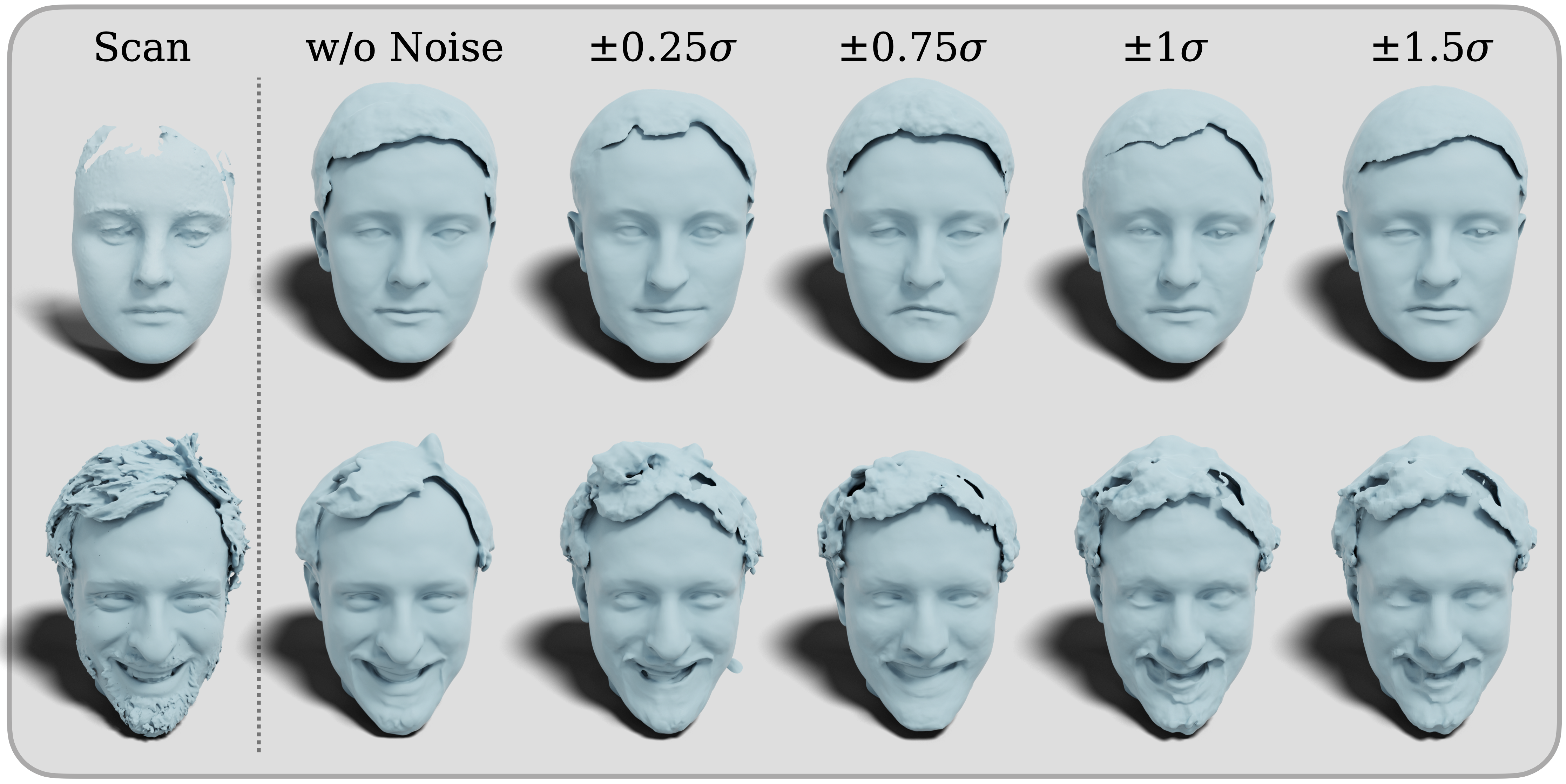}
    \captionof{figure}{
    \textbf{Qualitative Evaluation of fitting under Noisy Inputs}. We insert Gaussian noise to the input point clouds and measure the reconstruction performance. 
    } 
    \label{fig:noise}
\end{figure}
\section{Limitations and Societal Impact}
\begin{figure}[!h]
    \centering
    \includegraphics[width=\linewidth]{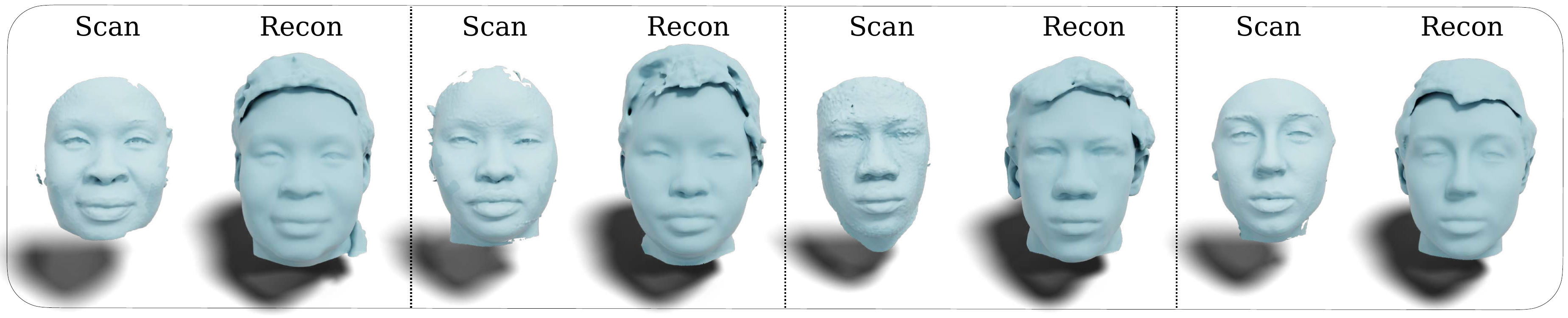}
    \caption{\textbf{Reconstruction performance on non-Caucasian ethnicities.} Despite the demographic biases, imHead can accurately reconstruct out-of-distribution samples.}
    \label{fig:bias}
\end{figure}

As stated in the main paper, although imHead makes a step towards full head modeling, it still suffers from some limitations. 
In particular, implicit models, in contrast to explicit 3DMMs suffer from slow inference times. 
To obtain a high resolution head it is required to sample and predict the SDF for a sufficient number of points which could significantly reduce the runtime of the method.
It must be also noted that SDFs require an additional post-processing marching cubes step which can further reduce the inference speed of the method. 
In contrast, 3DMMs can leverage fast rendering techniques and may provide a more efficient method in tasks where runtime performance is key priority. 
Implicit surfaces are also known to struggle capturing fine-grained details and fail to accurately model thin surfaces such as the hair strands. 
In addition, although as we experimentally show, imHead preserves the face correspondences there is not an 1-1 mapping similar to the case of explicit models. 
Furthermore, as noted in the main paper, localized editing is constrained by the fixed number of anchors that define each region. 
The editing process can also be influenced by the contributions of nearby local-part networks, which are designed to ensure smooth and plausible surfaces, but will affect the accuracy of edits especially at the boundaries.
Finally, despite curating a large-scale dataset, there are still race biases within the dataset. 
This also includes the hair regions which are directly adapted from the NPHM dataset, which has also limited diversity and cannot adequately represent all hair types. 
As an extend, imHead also shares the same demographic biases that should be taken into consideration when using imHead for downstreaming tasks. 
Despite the biases, as can be seen in \ref{fig:bias}, imHead can generalize well in out-of-distribution and non-Caucasian ethnicities. 

\section{Dataset Curation}
To enable large-scale head modeling we utilized MimicMe datasets~\cite{papaioannou2022mimicme}, which consists of 5,000 distinct subjects under different expressions. 
MimicMe dataset was collected using a 3dMD face capture system. 
The raw scans have a resolution of approximately 60,000 vertices. 
We filter the dataset to avoid noisy scans, resulting in a total of 4,000 distinct subjects being retained, with available metadata including gender ($57$\% male, $43$\% female), age ($1-81$ years old) and ethnicity ($73$\% White, $13$\% Asian, $7$\% Mixed and $3$\% Black, $4\%$ Other). Notably, the collected head scans demonstrate significant diversity across age, ethnicity, and height, marking progress toward a universal full head model. In comparison to previous implicit head models \cite{giebenhain2023nphm,zheng2022imface}, the curated dataset encompasses over 600 children under the age of 12, as well as more than 100 individuals aged over 60. 

\begin{figure}[!h]
    \centering
    \includegraphics[width=\linewidth]{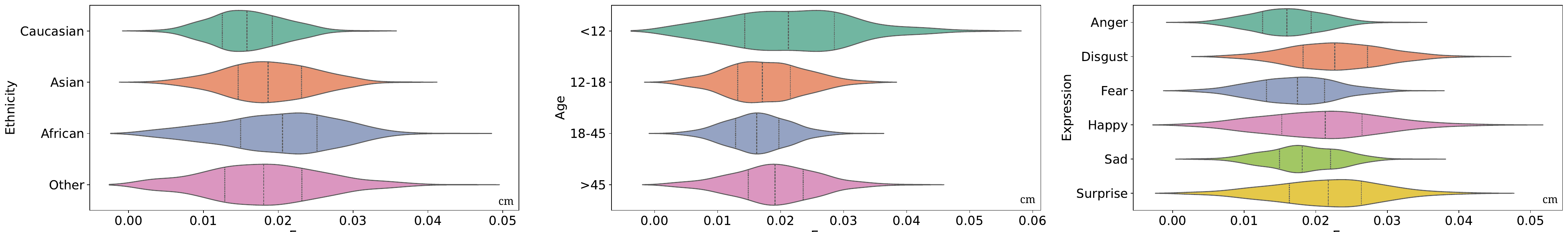}
    \caption{\textbf{Registration and Data Curation Errors.} We report reconstruction errors during the data curation process for different ethnicities and expressions.} 
    \label{fig:violin}
\end{figure}

To bring the raw scans into dense correspondence, we utilized a multi-step pipeline. Initially, the scans were rendered from multiple views and 2D joint locations were detected using RetinaFace \cite{deng2020retinaface}. Subsequently,  the 2D landmark locations were lifted to 3D by utilizing a linear triangulation and projected to the 3D surface.  Using the 3D detected keypoints, we fit FLAME parametric model by optimizing the pose and expression parameters to align the template head to the exact pose, expression and shape of each raw scan. 
Specifically, we optimize the pose $\boldmath{\theta}$, expression $\boldmath{\psi}$ and shape $\boldmath{\beta}$ parameters using following loss function: 
\begin{equation}
\mathcal{L} = \mathcal{L}_J + \mathcal{L}_{cd} +   ||\boldmath{\beta}||_2 + ||\boldmath{\psi}||_2  + ||\boldmath{\theta}||_2 
\end{equation}
where $\mathcal{L}_J = || J - \hat{J}||_2$ is a keypoint loss that enforces FLAME landmakrs $\hat{J}$ to match the detected keypoints $\hat{J}$ and $\mathcal{L}_{cd}$ is the chamfer distance loss that minimizes the scan to FLAME distance. The optimization process was performed using Adam optimizer with learning rate of $1e-3$.
We complete the full head of the aligned scans by fitting NPHM model~\cite{giebenhain2023nphm}.
However, a lot of the identity details of the subject might have been diminished during the fitting process. 
To retrieve the identity details we perform a Non-rigid Iterative Closest Point algorithm (NICP) \cite{amberg2007optimal} between the fitted meshes and the 3D raw scans. 
The proposed fitting and registration process enables the capture of rich facial details while ensuring plausible head surfaces with minimal reconstruction error. As shown in Fig.~\ref{fig:violin}, the non-rigid ICP step helps mitigate racial biases that may arise during the fitting process.

\section{Implementation Details}
In this section we provide the implementation details of the different components of our network. 
\subsection{Identity Network}
The identity network of the proposed imHead model is composed of three main modules: the \textit{Decomposition network}, the \textit{Local-Part Networks} and the \textit{Fusion network}. 
Bellow we describe the implementation details for each one of them: 

\noindent\textbf{Decomposition network.} The Decomposition network is responsible for the mapping of the global identity latent codes $\bm{z}_{id}$ to a set of localized embeddings $\{\bm{z}^{j}_{id}\}$ that span the 3D head.  
We define $\bm{z}_{id}$ using a simple Embedding layer that maps dataset instances to a 256-dimension latent code. 
Using a fully connected layer, we project the global latent $\bm{z}_{id}$ to $K$ embeddings of 32 dimensions each.
We follow NPHM~\cite{giebenhain2023nphm} and select $K=39$ keypoints that span the 3D head, resulting in a localized embedding with a total of 1248-dimensions. 
Using this simple yet efficient mapping we can achieve both compact global latent space, which can effectively improve the reconstruction capabilities of the network~\cite{tarasiou2024locally,Zheng_2024_CVPR} along with a localized intermediate representation that enables localized editing. 

\noindent\textbf{Landmark Regression.}
Following the latent embedding split, we use an MLP to regress the keypoints of the head, that will serve as the local coordinates for each region. 
In particular, we use a three-layer MLP that receives the set of local embeddings $\{\bm{z}^{j}_{id} \in \mathbb{R}^{32}\}$ as input and predicts the K=39 facial keypoints $\{\bm{k}^{j} \in \mathbb{R}^{3}\}$. 
We opted to use the intermediate local embedding representation to regress the facial landmarks as it can provide more robust estimations even after shape manipulations. 

\noindent\textbf{Local-Part Networks.}
Using a point sampled from the 3D space $\bm{x} \in \mathbb{R}^3$, we use an enseble of local-part networks to extract a point-specific feature $\mathbf{f}_j$ per region. 
To acquire the local part-specific feature $\mathbf{f}_j$, we feed point $\bm{x}$ along with the localized embeddings $\{\bm{z}^{j}_{id}\}$ to their corresponding local-part module. 
To better capture the high frequency details of the shapes \cite{potamias2022graphwalks}, we use a set of positional embeddings as defined in~\cite{mildenhall2020nerf}: 
\begin{equation*}
\begin{aligned}
    \gamma(\bm{x}) = \big(&\bm{x}, \\
    &\sin(2^0 \pi \bm{x}), \cos(2^0 \pi \bm{x}), \\
    &\sin(2^1 \pi \bm{x}), \cos(2^1 \pi \bm{x}), \ldots, \\
    &\sin(2^{L-1} \pi \bm{x}), \cos(2^{L-1} \pi \bm{x}) \big)
\end{aligned}    
\end{equation*}
that map the points $\bm{x}$ to a high dimensionality. 
We use $L=7$ frequency bands. 
Before feeding each point to the corresponding local-part network, we first normalize it according to the keypoint $\bm{k}^{j}_{id}$ associated with each part-network. 
This step is essential to normalize the coordinate system of each part network and not only achieve efficient and stable training but increase the expressivity of the network. 
We implement each local-part network using a small DeepSDF module with 4 layers and a hidden dimension~\cite{park2019deepsdf} of 200. 
Following the implementations of ~\cite{park2019deepsdf} we use \texttt{softplus} activation function. 

\noindent\textbf{Fusion Network.}
The final step of our identity network is to fuse the extracted feature codes $\mathbf{f}_j$ from each part-network $j$ back to a single global feature that will be used to regress the final SDF of point $\bm{x}$. 
Although an obvious choice would be to directly regress the fused SDF from the local-part networks, as we experimentally show in the ablation study, this choice significantly reduces the reconstruction quality and limits the editing properties of the network. 
We obtain the fused global feature vector using: 
\begin{equation}
    \hat{\mathbf{f}}_x = \sum_{j}^K {w}(\bm{x}, \bm{k}_j) \mathbf{f}^{j}_x
\end{equation}
where ${w}(\bm{x}, \bm{k}_j)$ scales the contribution of each feature embedding based on position of the point $\bm{x}$: 
\begin{equation}
{{w}}(\bm{x}, \bm{k}_j) = \frac{e^{\frac{-||\bm{x}- \bm{k}_j||_2}{\sigma}}}{\sum_{j}^K e^{\frac{-||\bm{x}- \bm{k}_j||_2}{\sigma}}}
\end{equation}
The final feature vector along with the correspond point $\bm{x}$ is then fed to the FusionNet to predict the final signed distance field $y$: 
\begin{equation}
    y = \mathcal{F}_{\theta}(\bm{x}, \hat{\mathbf{f}}_x) \in \mathbb{R}
\end{equation}
We implement the fusion network as a small DeepSDF module~\cite{park2019deepsdf} with 4 layers and 200 latent dimensions. 
Similar to the local-part networks, we use \texttt{softplus} activation function. 

\subsection{Expression Warping Module}
\begin{figure}[!ht]
    \centering
    \includegraphics[width=\linewidth]{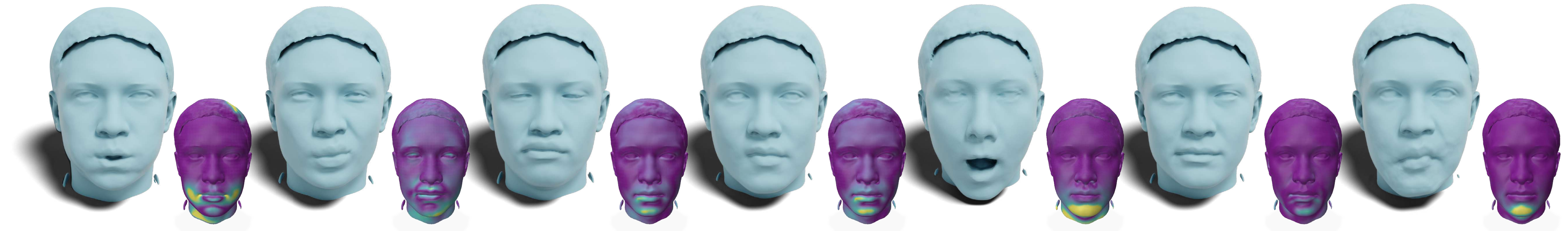} 
    \captionof{figure}{
    \textbf{Generalization to unseen expressions}. Given that imHead is relies on FLAME \cite{li2017learning} expression space, it can easily generate out-of-distribution expressions. 
    } 
    \label{fig:flame}
\end{figure}

Our expression module is responsible for backward-warping the sampled points from the observation space $\bm{x}_{obs} \in \mathbb{R}^3$ to the canonical space of the identity network. 
To enable fast integration to existing pipelines we define $\bm{z}_{exp}$ using the expression parameters of FLAME model~\cite{li2017learning} acquired during the fitting process of the dataset.
The FLAME expressions are then fed to a higher dimensional latent space and used to condition the expression warping module. 
Given that imHead is conditioned on FLAME expression parameters, it can be easily adapted to existing pipelines and generalize to unseen expressions as shown in \cref{fig:flame}
Similar to the previous networks, we implement the expression module using a DeepSDF network with 8-layers with 128-hidden dimensions.

\section{Backward vs. Forward Warping}
\begin{figure}[!ht]
    \centering
     \includegraphics[width=\linewidth]{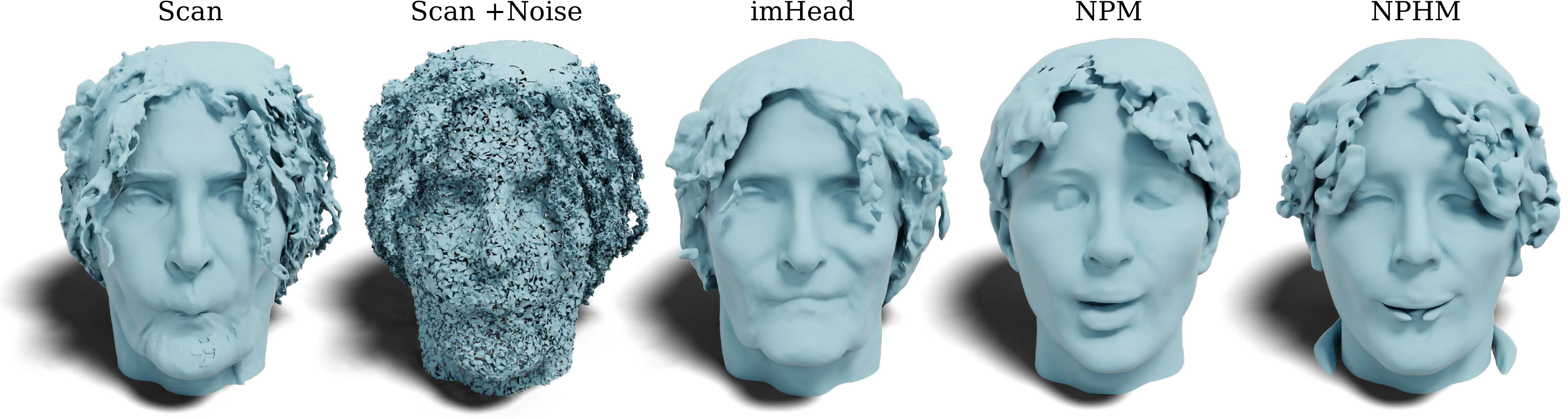} 
    \captionof{figure}{
    \textbf{Failure cases of forward deformation methods}. Given that forward warping methods rely on iterative root-finding schemes, inaccurate correspondences can significantly impact reconstruction performance. 
    } 
    \label{fig:fail}
\end{figure}

Backward warping has been widely used across implicit field \cite{Zhao_2022_CVPR,zheng2022imface,athar2022rignerf} achieving robust results and offering several advancements over traditional forward deformation warping. 
Specifically, backward warping does not require any costly registration process to bring the scans in dense correspondence. 
In contrast, forward deformation methods such as NPM~\cite{giebenhain2023nphm} and NPHM~\cite{giebenhain2023nphm} require a registration step to non-rigidly aling the scans to calculate the target deformation fields. 
Additionally, forward deformation methods heavily rely on iterative root finding schemes, which apart from time consuming optimization processes introduced, can also affect the robustness of the parametric model. 
In particular, as shown in \cref{fig:fail}, forward deformation methods, can fail in cases of noisy scans where the inverse correspondences are not established correctly



{
    \small
    \bibliographystyle{ieeenat_fullname}
    \bibliography{main}
} 

\end{document}